\documentclass[conference]{IEEEtran}
\IEEEoverridecommandlockouts
\usepackage{cite}
\usepackage{amsmath,amssymb,amsfonts}
\usepackage[numbers]{natbib}
\usepackage[colorlinks,citecolor=black,linkcolor=black]{hyperref}
\usepackage{marvosym}
\usepackage{pifont}
\usepackage{algorithmic}
\usepackage{graphicx}
\usepackage{subfig}
\usepackage{textcomp}
\usepackage{xcolor}
\def\BibTeX{{\rm B\kern-.05em{\sc i\kern-.025em b}\kern-.08em
    T\kern-.1667em\lower.7ex\hbox{E}\kern-.125emX}}
\begin{document}

\title{ESMC: Entire Space Multi-Task Model for Post-Click Conversion Rate via Parameter Constraint}

\author{
\IEEEauthorblockN{
Zhenhao Jiang\IEEEauthorrefmark{4}\IEEEauthorrefmark{3}\IEEEauthorrefmark{1}\thanks{This paper was done when Zhenhao Jiang as an intern in Alibaba Group.}, 
Biao Zeng\IEEEauthorrefmark{2}\IEEEauthorrefmark{1}\thanks{\IEEEauthorrefmark{1} Both authors contributed equally to this paper.},
Hao Feng\IEEEauthorrefmark{2},
Jin Liu\IEEEauthorrefmark{2}\textsuperscript{\Letter},
Jicong Fan\IEEEauthorrefmark{4}\IEEEauthorrefmark{3}\textsuperscript{\Letter},\\ 
Jie Zhang\IEEEauthorrefmark{2}\thanks{\textsuperscript{\Letter} Corresponding authors.},
Jia Jia\IEEEauthorrefmark{2}, 
Ning Hu\IEEEauthorrefmark{2},
Xingyu Chen\IEEEauthorrefmark{5},
Xuguang Lan\IEEEauthorrefmark{5}}
\IEEEauthorblockA{\IEEEauthorrefmark{4}School of Data Science, The Chinese University of Hongkong, Shenzhen, China}
\IEEEauthorblockA{\IEEEauthorrefmark{2}Alibaba Group, Shanghai\&Hangzhou, China}
\IEEEauthorblockA{\IEEEauthorrefmark{3}Shenzhen Research Institute of Big Data, Shenzhen, China}
\IEEEauthorblockA{\IEEEauthorrefmark{5}Xi'an Jiaotong University, Xi'an, China}
\IEEEauthorblockA{222041010@link.cuhk.edu.cn, \{biaozeng.zb, zhisu.fh, nanjia.lj\}@alibaba-inc.com, fanjicong@cuhk.edu.cn} }

\maketitle

\begin{abstract}
Large-scale online recommender system spreads all over the Internet being in charge of two basic tasks: Click-Through Rate (CTR) and Post-Click Conversion Rate (CVR) estimations. However, traditional CVR estimators suffer from well-known Sample Selection Bias and Data Sparsity issues. 
Entire space models were proposed to address the two issues via tracing the decision-making path of ``exposure\_click\_purchase". Further, some researchers observed that there are purchase-related behaviors between click and purchase, which can better draw the user's decision-making intention and improve the recommendation performance. Thus, the decision-making path has been extended to ``exposure\_click\_in-shop action\_purchase" and can be modeled with conditional probability approach. Nevertheless, we observe that the chain rule of conditional probability does not always hold. We report Probability Space Confusion (PSC) issue and give a derivation of difference between ground-truth and estimation mathematically. We propose a novel Entire Space Multi-Task Model for Post-Click Conversion Rate via Parameter Constraint (ESMC) and two alternatives: Entire Space Multi-Task Model with Siamese Network (ESMS) and Entire Space Multi-Task Model in Global Domain (ESMG) to address the PSC issue. Specifically, we handle ``exposure\_click\_in-shop action" and ``in-shop action\_purchase" separately in the light of characteristics of in-shop action. The first path is still treated with conditional probability while the second one is treated with parameter constraint strategy. Experiments on both offline and online environments in a large-scale recommendation system illustrate the superiority of our proposed methods over state-of-the-art models. The code and real-world datasets will be released for further research.
\end{abstract}


\section{Introduction}
Selecting best-suited products from floods of candidates to deliver them to users based on their appropriate  preferences has become a significant task on most of online platforms such as online food booking, short video, e-commerce, \textit{etc} \cite{zhou2018deep, lin2022feature, xu2022amcad, zhang2022picasso}. Recommender system plays an important role to handle the task timely and accurately with the help of deep learning algorithms \cite{zhang2022towards, wang2023sequential}. Recommendation service first recalls candidates from item pool and then feeds them into a recommender algorithm to predict several metrics such as Click-Through Rate (CTR) and Post-Click Conversion Rate (CVR) \cite{kumar2015predicting, richardson2007predicting, lee2012estimating}. Next, items are ranked according to CTR, CVR or other metrics and exposed on the terminal device of user. A user may click an item to enter the in-shop page to add it into cart/wish-list and purchase, which can be described as a decision-making graph of ``exposure\_click\_in-shop action\_purchase" \cite{takada2021pop, gupte2020automated}. This feedback will be recorded and used for updating recommender algorithm to ensure that system can capture the evolution of interest and recent preference of the user. To provide users with more accurate recommendation service, a high-quality CVR estimator is crucial in practice \cite{zhou2019deep}. 

\begin{figure*}[h]
    \centering
\subfloat[Illustration of SSB and DS issues in CVR estimation that model is trained over clicked samples while is used to infer on exposed samples. The size of samples diminishes from exposure to purchase.]{\includegraphics[width=.45\linewidth]{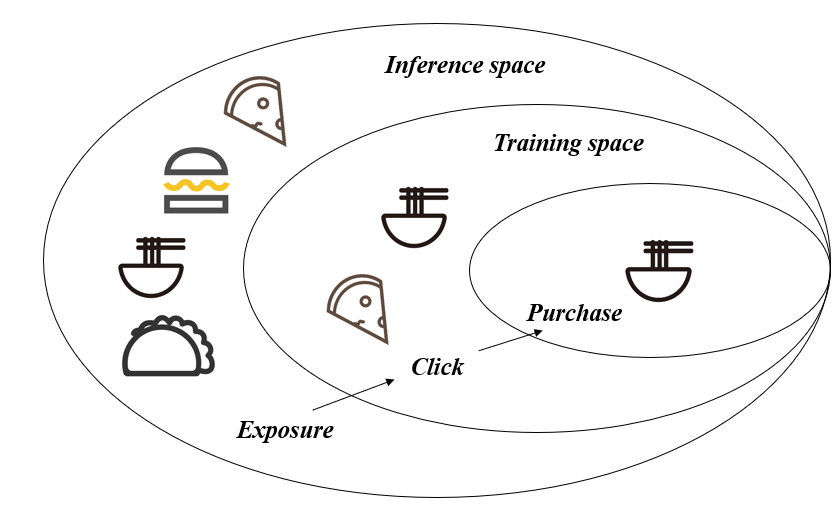} \label{fig2a}}\hspace{2pt}
\subfloat[Demonstration of Probability Space Confusion issue. Cart and Purchase may not be in the same visit.]
 {\includegraphics[width=.45\linewidth]{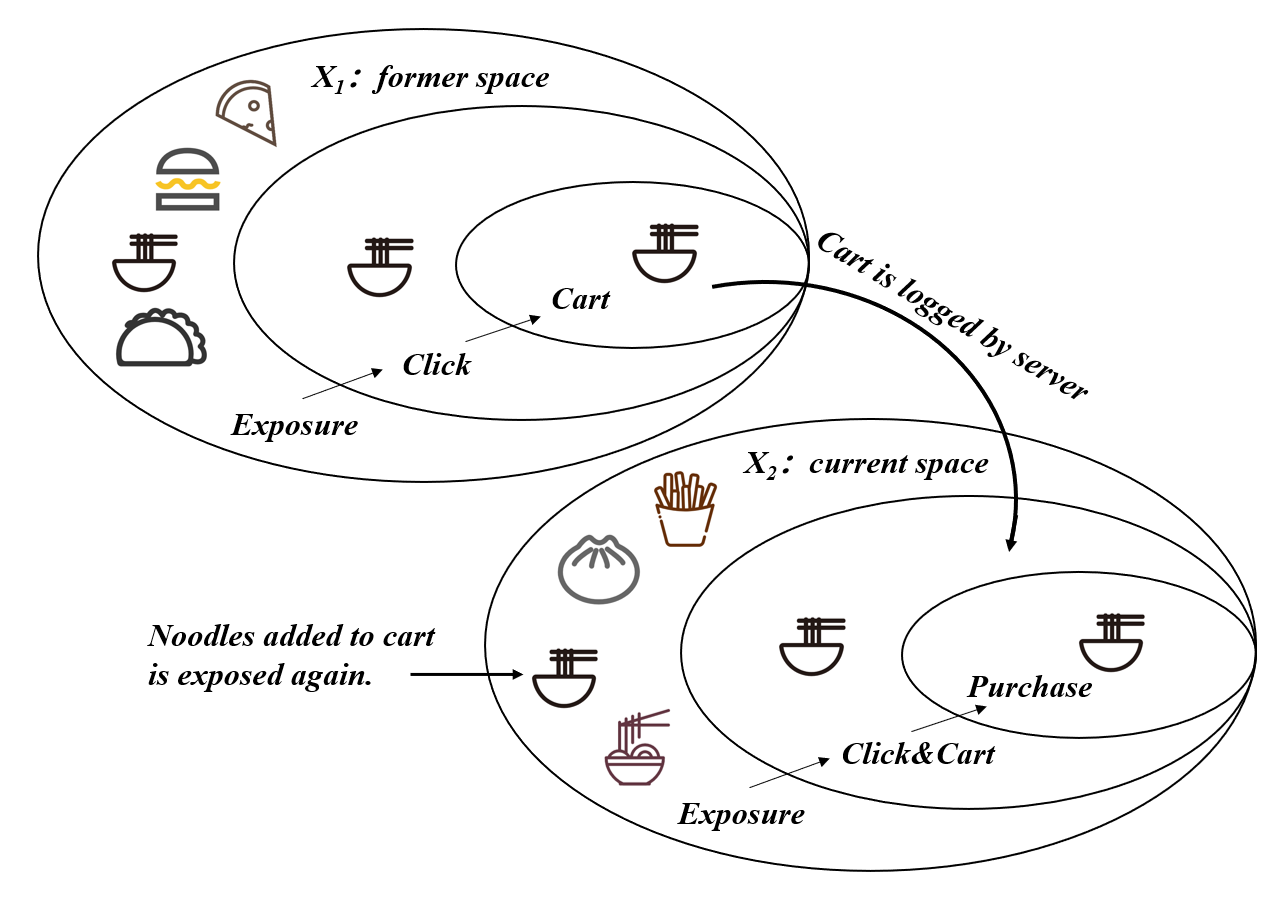} \label{fig2b}}\hspace{2pt}
 \\
 \subfloat[Demonstration of our approach for addressing the PSC issue. The orange line means sample space calibration and the blue line means information injection.]
 {\includegraphics[width=.9\linewidth]{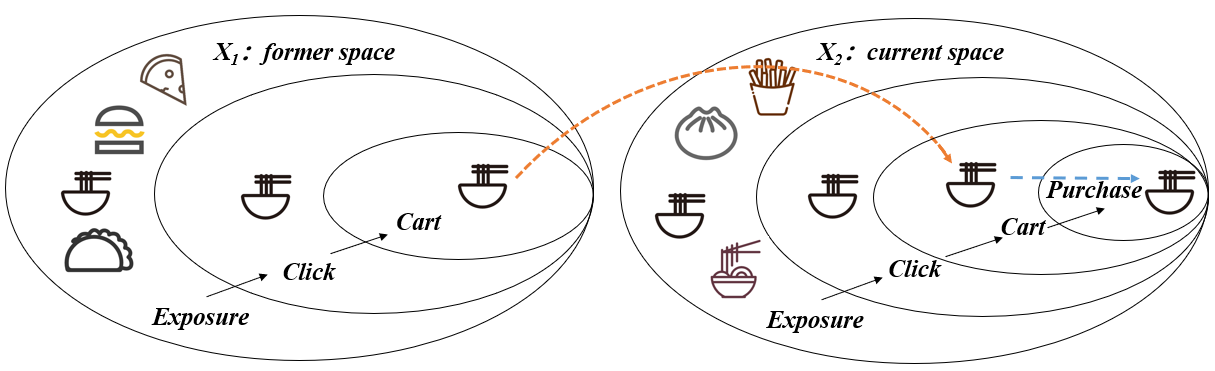} \label{fig2c}}

\caption{Illustration of sample space. (a) SSB and DS issues. (b) Key problem in this paper: PSC issue. (c) Key idea in this paper to handle PSC issue.}
\label{fig1}
\end{figure*}

There are two basic issues in the CVR estimation task: Sample Selection Bias (SSB) and Data Sparsity (DS) \cite{ma2018entire} shown in Fig. \ref{fig2a}. SSB refers to a gap between training sample space and online inference sample space. Traditional CVR estimators are trained on clicked samples while implemented on exposed samples based on the schema of online recommendation service. DS refers to the issue that the size of clicked samples is too small to train a model that can fit conversion well. Consequently, the performance of recommender algorithm is dissatisfactory in online service \cite{wang2022escm2}. SSB and DS are fundamental issues that we must overcome in industrial recommender systems. Many researchers have proposed entire space models to address SSB and DS \cite{wen2021hierarchically, gao2023rec4ad, liu2022rating}. Entire Space Multi-Task Model (ESMM) is one of the representatives of entire space models that will be presented in Section 3. Following ESMM, Entire Space Multi-Task Model via Behavior Decomposition (ESMM\textsuperscript{2}) is proposed to introduce in-shop behaviors to estimate CVR with the almost same ideology of ESMM \cite{wen2020entire}.

Several studies have claimed that user may purchase items from shopping cart or wish list and we observe this phenomenon in real business as well \cite{peska2011upcomp, zhang2022price, pradhan2012wish}. The action of adding to cart/wish-list (in-shop action\footnote{In this paper, we focus on adding to cart.}) bridges click and purchase that is more conversion-related than click.  Therefore, extracting the functionality of in-shop action in decision-making path is meaningful. In ESMM\textsuperscript{2}, algorithm explicitly models the sequential behavior of ``exposure\_click\_in-shop action\_purchase" via conditional probability to leverage samples over the entire space to address SSB and DS issues more efficiently. However, the probability-based approach doesn't always work. In-shop action may come from other sample space not contained in the current exposure space different from click or purchase.

In this paper, we report \textbf{Probability Space Confusion (PSC)} problem of ESMM\textsuperscript{2}-like models\footnote{In this paper, ESMM\textsuperscript{2}-like model is defined as the model considering both decision-making graph of ``exposure\_click\_in-shop action\_purchase" and the probabilistic dependence among different behaviors.} shown in Fig.~\ref{fig2b} which will be presented in detail in Section 4. \textbf{Because ESMM\textsuperscript{2}-like model is widely used in industrial recommenders, it is critical and meaningful to improve it.} We also present the derivation of the gap between estimation and ground-truth mathematically under PSC issue and propose Entire Space Multi-Task Model via Parameter Constraint (ESMC) that mainly consists of three modules: 1) shared embedding (SE), 2) constrained twin towers (CTT), and 3) sequential composition module (SCM) and one strategy: Sample Calibration to address PSC problem. Before training, Sample Calibration unifies the sample space. In the model, SE maps feature vectors into low-dimensionally dense vectors at first. Then, CCT fits Click-Through Conversion Rate (CTCVR) and Click-Through Cart Adding Rate (CTCAR) under a given constraint. Finally, SCM combines CTR, CTCAR and CTCVR together to perform a multi-task estimation. Going further, we present two alternatives (\textit{i.e.} ESMS and ESMG) and discuss their advantages and disadvantages to help practitioners choose the most suitable solution for their own business.

The main contributions of this work are as follows:
\begin{itemize}
    \item This is the first work that reports the PSC issue in CVR estimation with in-shop behaviors. We demonstrate the problem from the perspective of sample space and emphasize the importance of distinguishing between click/purchase and in-shop actions. We also highlight the mathematical theory behind the PSC issue.
    \item We propose ESMC, the first work that enhances ESMM\textsuperscript{2} with a novel parameter constraint approach. ESMC avoids the PSC issue and improves the performance of ESMM\textsuperscript{2}. Extensive experimental results verify our claims.
    \item We also propose two alternatives of ESMC (\textit{i.e.} ESMS and ESMG) and discuss their characteristics to help others identify the most suitable strategy to address the PSC issue in their own business. 
    \item To support future research, we construct real-world datasets collected from a large-scale online food platform, which we will release publicly. 
\end{itemize}

The important abbreviations in this paper are summarized in Table~\ref{abb}.

\begin{table}[t]
    \centering
    \caption{Summarization of important abbreviations.}
    \begin{tabular}{cc}
    \hline
      Abbreviation   & Description \\
    \hline
    Cart & the behavior of adding to cart\\
      CTR & Click-Through Rate\\
      CVR & Post-Click Conversion Rate\\
      CAR & Cart Adding Rate\\
      CTCAR & Click-Through Cart Adding Rate\\
      CTCVR & Click-Through Conversion Rate\\
      PSC & Probability Space Confusion\\
      SSB & Sample Selection Bias\\
      DS & Data Sparsity\\
    \hline
    \end{tabular}
    \label{abb}
\end{table}

\section{Related Works}
\subsection{Multi-Task Learning}
Since it is necessary to estimate multiple tasks (\textit{i.e.} CTR and CVR) simultaneously in recommendation system, it is critical to design a multi-task learning model. Deep recurrent neural network is employed to encode the text sequence into a latent vector, specifically gated recurrent units trained end-to-end on the collaborative filtering task \cite{bansal2016ask}. MMoE consists of multiple expert networks and gate networks to learn the correlations and differences among different tasks to fit multiple downstream tasks \cite{ma2018modeling}. The two basic tasks in recommendation (\textit{i.e.} rank and rate) are traced simultaneously with a multi-task framework in \cite{hadash2018rank}. NMTR considers the underlying relationship among different types of behaviors and performs a joint optimization with a multi-task learning strategy, where the optimization on a behavior is treated as a task \cite{gao2019neural}. In \cite{lu2018like}, a multi-task recommendation model with matrix factorization is proposed which jointly learns to give rating estimation and recommendation explanation. MTRec is designed based on heterogeneous information network equipped with a Bayesian task weight learner that is able to balance two tasks during optimization automatically and provide a good interpretability \cite{li2020multi}. SoNeuMF is an extension of neural matrix factorization and is able to simultaneously model the social domain and item domain interactions via sharing user representation in two tasks \cite{feng2022social}. AMT-IRE is a multi-task framework which can adaptively extract the inner relations between group members and obtain consensus group preferences with the help of attention mechanism \cite{chen2021attentive}. PLE is used to solve the problem of negative transfer in multi-task learning. It can be considered as a stacked structure of basic modules of MMoE, introducing specific expert networks and common expert networks to decouple different tasks \cite{tang2020progressive}. 

\subsection{Conversion Rate Prediction}
ESMM proposes a decision-making path of ``exposure\_click\_purchase" and draws CVR based on the chain rule of conditional probability \cite{ma2018entire}. ESMM\textsuperscript{2} extends the decision-making path to ``exposure\_click\_in-shop action\_purchase" with the similar idea of ESMM \cite{wen2020entire}. In \cite{zhang2020large}, researchers find that CVR estimation in basic ESMM is biased and address this problem with causal approach (ESCM). ESCM\textsuperscript{2} gives out a more solid proof for the bias issue in ESMM and employs a similar solution in ESCM \cite{wang2022escm2}. Here, we first report a novel PSC issue in ESMM\textsuperscript{2} and provide three solutions to address it. There are also many studies that predict CVR from other perspectives. In \cite{lee2012estimating}, researchers model CVR at different hierarchical levels with separate binomial distributions and estimate the distribution parameters individually. ACN uses Transformer to implement feature cross-over and employs capsule networks with a modified dynamic routing algorithm integrating with an attention mechanism to capture multiple interests from user behavior sequence \cite{li2021attentive}. GCI counterfactually predicting the probability of each specific group of each unit belongs to for post-click conversion estimation \cite{gu2021estimating}. AutoHERI leverages the interplay across multi-tasks' representation learning. It is designed to learn optimal connections between layer-wise representations of different tasks and can be easily extended to new scenarios with one-shot search algorithm \cite{wei2021autoheri}.  

Unlike the studies mentioned above, we focus on the distinctiveness of Cart from a probability perspective. Our emphasis is on explaining the mathematical theory behind it and proposing simple yet effective solutions.

\section{Preliminary}
Since this paper aims to improve ESMM\textsuperscript{2}, we first introduce ESMM and ESMM\textsuperscript{2} in this Section. 

\begin{figure*}[t]
	\centering
	\subfloat[ ]{\includegraphics[width=.65\columnwidth]{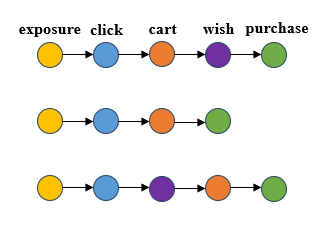}}\hspace{5pt}
	\subfloat[ ]{\includegraphics[width=.65\columnwidth]{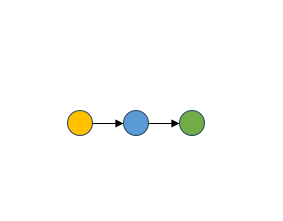}}\hspace{5pt}
        \subfloat[ ]{\includegraphics[width=.65\columnwidth]{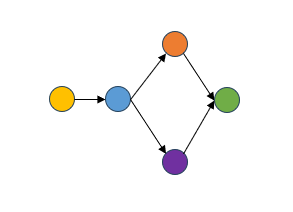}}
 \caption{Illustration of three types of user decision graph from exposure to purchase. (a) Three real decision-making graphs online. (b) Decision-making graph in ESMM. (c) Decision-making graph in ESMM\textsuperscript{2}.}
 \label{fig3}
\end{figure*}

\subsection{Problem Formulation}

\begin{table}[t]
\centering
\caption{Summarization of notations.}
\label{notation}
\begin{tabular}{cc}
\hline
   Notation  & Description \\
\hline
$u$ & user\\
   $\mathcal{V}$  & items exposed to user\\
   $v$ & item\\
   $\mathcal{C}/c_{u,v}$ & clicked items by user/entry of $\mathcal{C}$\\
   $\mathcal{O}/o_{u,v}$ & purchased items by user/entry of $\mathcal{O}$\\
   $\mathcal{A}/a_{u,v}$ & items added in cart by user/entry of $\mathcal{A}$\\
   $\mathbb{X}$ & exposure space\\
   $\mathbb{C}$ & click space\\
   $\mathbb{O}$ & conversion space\\
   $\mathbb{A}$ & Cart space\\
   $X/x$ & exposure event/value of $X$\\
   $C/c$ & click event/value of $C$\\
   $A/a$ & click \& Cart event/value of $A$\\
   $R/r$ & conversion event/value of $R$\\
   letters with hat (\textit{e.g.} $\hat{r}$) & the corresponding estimators given by algorithm\\
\hline
\end{tabular}
\end{table}

Here, we state the Post-Click Conversion Rate estimation problem on the
entire space with Cart. Let $u$ denote a user browsing item feeds and item set $\mathcal{V}=\{v_1, v_2, \dots, v_m\}$ represent the items on exposure space $\mathbb{X}$ for $u$. Define $\mathcal{C}$ as the click set that indicates which item in $\mathcal{V}$ is clicked by $u$ where each entry $c_{u,v} \in \{0, 1\}$, $\mathcal{O}$ as the conversion (purchase) set that indicates which item in $\mathcal{V}$ is conversed finally where each entry $o_{u,v} \in \{0, 1\}$. $\mathbb{C}$ and $\mathbb{O}$ indicates the click space and conversion space, respectively. Especially, let $\mathcal{A}$ denotes the collection of items added in cart where each entry $a_{u,v} \in \{0, 1\}$ and $\mathbb{A}$ is the Cart space. The notations used in this paper are summarized in Table~\ref{notation}.

In practice, online recommender server has to estimate CTR and CVR on the exposure space $\mathbb{X}$. Consequently, we have to train a model in this manner to keep the online-offline consistency (avoiding sample selection bias). Further, if $O$ is fully observed, the ideal loss function is formulated as:
\begin{equation}
    L:=\mathbb{E}_{u,v}[\delta(o_{u,v},\hat{o}_{u,v})],
\end{equation}
where $\mathbb{E}$ means expectation of events, $\hat{o}_{u,v}$ is estimated result, and $\delta$ is an error function such as the cross entropy loss:
\begin{equation}
    \delta(o_{u,v},\hat{o}_{u,v}):=-o_{u,v}\log \hat{o}_{u,v}-(1-o_{u,v})\log (1-\hat{o}_{u,v}).
    \label{eq2}
\end{equation}

\subsection{Entire Space Multi-Task Model}
On online shopping platform, an item might experience ``exposure\_click\_purchase" to converse. In the light of this process, ESMM proposes a CVR estimation approach via chain rule \cite{ma2018entire}:
\begin{equation}
    \mathbb{P}(CTCVR)=\mathbb{P}(CTR)\times\mathbb{P}(CVR),
\end{equation}
CTCVR estimation is given out with the product of CTR and CVR predicted by two full-connected towers. During the training process, ESMM minimizes the empirical risk of
CTR and CTCVR estimation over $\mathbb{X}$:
\begin{equation}
    \begin{split}
    L_{CTR}&=\mathbb{E}_{u,v}[\delta(c_{u,v}, \hat{c}_{u,v})]\\L_{CTCVR}&=\mathbb{E}_{u,v}[\delta(c_{u,v}\times o_{u,v}, \hat{c}_{u,v}\times\hat{o}_{u,v})].
    \end{split}
\end{equation}

Thus, ESMM addresses SSB problem via training on the exposure space. Additionally, since the size of clicked samples is much larger than that of conversion samples, modeling CVR on the exposure space allows for better utilization of the available data to tackle DS problem.

\subsection{Entire Space Multi-Task Model via Behavior Decomposition }
This is an extension work of ESMM, known as ESMM\textsuperscript{2} \cite{wen2020entire}. The basic ideology of them are significantly similar. Compared with ESMM, the main improvement is that ESMM\textsuperscript{2} involves intermediate behaviors between click and purchase, such as Cart and ``adding to wish-list" and is more in line with the real decision-making process in online service for user. Actually, different actions cannot be triggered at the same time. To simplify the problem, ESMM\textsuperscript{2} considers that all in-shop actions can be triggered in parallel. For the sake of description, we focus on Cart.

Similar to ESMM, ESMM\textsuperscript{2} employs chain rule to model CVR with behaviors:
\begin{equation}
    \mathbb{P}(CTCVR)=\mathbb{P}(CTR)\times\mathbb{P}(CAR)\times\mathbb{P}(CVR),
    \label{eq6}
\end{equation}
where CAR is Cart Adding Rate on $\mathbb{C}$ and CVR is Conversion Rate on $\mathbb{A}$. Further, the probability of CTCVR of item $v$ can be defined as the conditional probability mathematically in accordance with ``exposure\_click\_Cart\_purchase" process:
\begin{equation}
\begin{split}
    \mathbb{P}(o_{u,v}=1|c_{u,v}=1)=\\
    \mathbb{P}(o_{u,v}=1|c_{u,v}=1,a_{u,v}=1)\mathbb{P}(a_{u,v}=1|c_{u,v}=1).
\end{split}
\end{equation}
Undoubtedly, an item cannot be added to the cart without being clicked, and it cannot be purchased without being added to the cart. Therefore, CTCVR can be modeled in a similar way to other in-shop behaviors. The decision-making path is illustrated in Figure \ref{fig3}.

However, does the chain rule always hold? 

\section{Discussion on ESMM\textsuperscript{2}}
In this section, we first explain the PSC issue. We then provide a mathematical derivation for quantifying the gap between the ground-truth and estimated values. Finally, we discuss the implications of the gap.

\subsection{Probability Space Confusion Issue}
\begin{figure*}
    \centering
    \includegraphics[width=.95\linewidth]{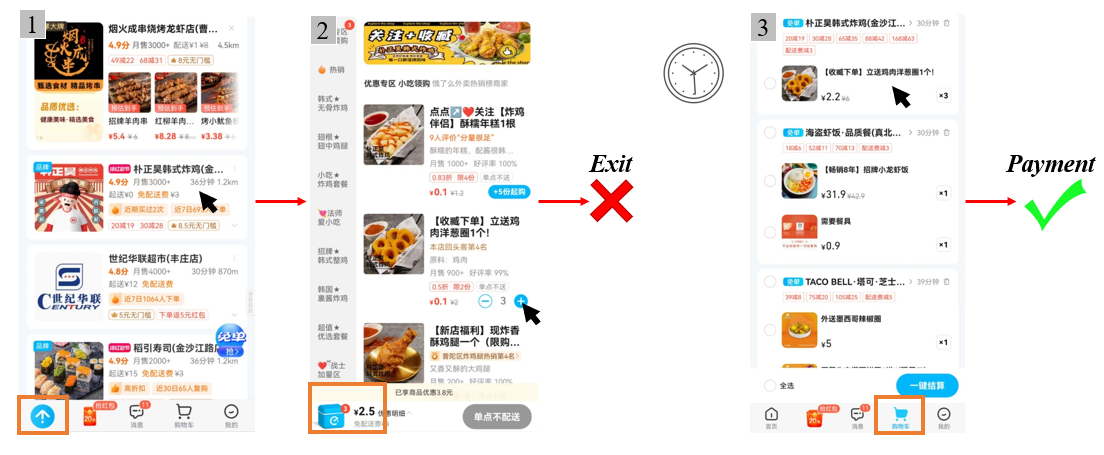}
    \caption{Illustration of Eleme APP, an Alibaba Group's takeaway platform that serves hundreds of millions of users. The black arrow represents the user's behavior. Page 1 (shown at the top left corner) is the recommendation page, page 2 is the detail page (in-shop page), and page 3 is the shopping cart page. A user may click on an item in the recommendation page to enter the in-shop page and add a product to shopping cart. And then, the user exits the platform. After a while, the user logs onto the platform again and goes directly to the shopping cart page to purchase the product. Therefore, the decision-making path of this user does not occur within the same visit.}
    \label{fig4}
\end{figure*}

ESMM\textsuperscript{2} introduces in-shop actions to draw the fine-grained decision-making process. It considers CVR on the exposure space and $\mathbb{C}, \mathbb{A},$ and $\mathbb{O}$ are sub-space defined on $\mathbb{X}$. When a user opens the online recommendation feeds, several items are exposed where the user can see. Then, the user may click one of the items to enter the detail page (in-shop page), add products to cart, and make the final payment. ESMM\textsuperscript{2} assumes the actions in the path of ``exposure\_click\_Cart\_purchase" occur within the same visit or in the same sample space. However, this assumption does not always hold true on real online platforms, as shown in Figure \ref{fig4}. The user may exit the detail page without making an immediate purchase after adding products to cart. Most online shopping platform records users' Cart information so that users can quickly find the items they prefer. Therefore, the user may log on the online platform again after a period of time and enter the shopping cart to buy. As a result, the paths of ``exposure\_click\_Cart" and ``Cart\_purchase" are in different visits or in different sample spaces. This raises a problem. Based on the assumption of ESMM\textsuperscript{2}, the entire space or exposure space is actually defined on the sample space of user's current visit. Because the user's information will be updated according to their behavior in the next visit, the recommender could predict the recommendation lists only based on the current status of users and items. Therefore, the exposure space for each visit is actually independent for a user. The calculation of probabilities defined on different sample spaces leads to the PSC issue. \textbf{Remark:} To simplify the problem, we use Session \footnote{For a web address, one session is equivalent to one visit.} \cite{wang2021survey} to determine whether actions occur within the same visit.

\subsection{Mathematical Derivation on PSC Issue}
Here we provide a mathematical derivation to evaluate the gap between the ground-truth and estimation of ESMM\textsuperscript{2}.

First, we give out the right expectation of $R$ under the case that the item has been already added to cart before. Based on the discussion of the PSC issue, the entire path are not in the same sample space in this case, defined as Bad Case. Additionally, $X_1$ is for the former exposure space while $X_2$ is for the current one.
\begin{equation}
    \begin{split}
        \mathbb{E}_{X_2}[R]&\overset{\textcircled{1}}{=}\mathbb{E}_{X_2}[R|A]\cdot\mathbb{E}_{X_1}\left[\frac{A}{C}\right]\\
        &\overset{\textcircled{2}}{=}\mathbb{E}_{X_2}[R|A]\cdot\int_{X_1}\frac{a}{c}\mathbb{P}(a,c)d(a,c)\\
        &=\mathbb{E}_{X_2}[R|A]\cdot\int_{X_1}a\mathbb{P}(a)da\cdot\int_{X_1}\frac{1}{c}\mathbb{P}(c)dc\\
        &=\mathbb{E}_{X_2}[R|A]\cdot\mathbb{E}_{X_1}[A]\cdot\mathbb{E}_{X_1}\left[\frac{1}{C}\right].
    \end{split}
\end{equation}

In the anticipation of ESMM\textsuperscript{2}, in-shop action always satisfies the chain rule of probability. Thus the expectation of estimator $\hat{R}$ given by ESMM\textsuperscript{2} in the Bad Case is:
\begin{equation}
    \begin{split}
        \mathbb{E}_{X_2}[\hat{R}]&\overset{\textcircled{3}}{=}\mathbb{E}_{X_2}\left[\frac{\hat{R}}{\hat{A}}\right]\cdot\mathbb{E}_{X_2}\left[\frac{\hat{A}}{\hat{C}}\right]\\
        &=\int_{X_2}\frac{\hat{r}}{\hat{a}}\mathbb{P}(\hat{r},\hat{a})d(\hat{r},\hat{a})\int_{X_2}\frac{\hat{a}}{\hat{c}}\mathbb{P}(\hat{a},\hat{c})d(\hat{a},\hat{c})\\
        &=\mathbb{E}_{X_2}\left[\frac{1}{\hat{C}}\right]\cdot\mathbb{E}_{X_2}[\hat{R}]\cdot\mathbb{E}_{X_2}\left[\frac{1}{\hat{A}}\right]\cdot\mathbb{E}_{X_2}[\hat{A}]\\
        &=\mathbb{E}_{X_2}[\hat{R}]\cdot\mathbb{E}_{X_2}\left[\frac{1}{\hat{A}}\right]\cdot\mathbb{E}_{X_2}[\hat{A}]\cdot\mathbb{E}_{X_2}\left[\frac{1}{\hat{C}}\right].
    \end{split}
\end{equation}

There are some explanations of the above derivation:

\textcircled{1} means that the user clicked on the item to enter the detail page and added products to cart in the previous visit. In current visit the user purchase products in cart. Because the shopping cart generally exists independently on the online platform (\textit{i.e.} not in the recommendation feeds), the click behavior of the current visit is not considered and the behavior in the current visit is expressed in terms of conditional probability here.

\textcircled{2} holds under the assumption $A\perp C$.

\textcircled{3} holds because ESMM\textsuperscript{2} considers that the entire path is in the same visit and satisfy the chain rule. It doesn't take into account the specificity of Cart.

Consider the gap between group-truth and estimation in the Bad Case:
\begin{equation}
    \begin{split}
        \text{Gap}
        &=\mathbb{E}_{X_2}[R]-\mathbb{E}_{X_2}[\hat{R}] \\
        &=\mathbb{E}_{X_2}[R|A]\cdot\mathbb{E}_{X_1}[A]\cdot\mathbb{E}_{X_1}\left[\frac{1}{C}\right]\\
        &-\mathbb{E}_{X_2}[\hat{R}]\cdot\mathbb{E}_{X_2}\left[\frac{1}{\hat{A}}\right]\cdot\mathbb{E}_{X_2}[\hat{A}]\cdot\mathbb{E}_{X_2}\left[\frac{1}{\hat{C}}\right].
    \end{split}
    \label{eq11}
\end{equation}

Here, we define $\mathbb{E}_{X_2}[R|A]$ and $\mathbb{E}_{X_2}[\hat{R}]\cdot\mathbb{E}_{X_2}\left[\frac{1}{\hat{A}}\right]$ as Left Terms, and $\mathbb{E}_{X_1}[A]\cdot\mathbb{E}_{X_1}\left[\frac{1}{C}\right]$ and $\mathbb{E}_{X_2}[\hat{A}]\cdot\mathbb{E}_{X_2}\left[\frac{1}{\hat{C}}\right]$ as Right Terms.

If this gap can be eliminated, then there must be an upper bound, so we derive a loose upper bound to prove the solvability of this problem.
\begin{equation}
\begin{split}
    \text{Gap}&=\int_{X_2}r\mathbb{P}(r|a)dr\int_{X_1}\frac{a}{c}\mathbb{P}(a,c)d(a,c)\\
    &-\int_{X_2}\frac{r}{a}\mathbb{P}(r,a)d(r,a)\int_{X_2}\frac{a}{c}\mathbb{P}(a,c)d(a,c)\\
    &\overset{\textcircled{4}}{\leq} \int_{X_2}rdr\int_{X_1}\frac{a}{c}d(a,c).
\end{split}
\end{equation}

\textcircled{4} holds because the range of arbitrary probability is $[0, 1]$.

It is evident that the integral domain is a finite interval and the integrand is bounded on the integral domain. There is always a finite upper bound to the gap, and therefore the problem is solvable.

\subsection{Discussion on The Difference}
In \eqref{eq11}, compared with the ground-truth, we can find that there are two differences in terms of formula form. 
\begin{itemize}
    \item \textbf{Left Terms.} The Cart information and the purchase information are decoupled in the current space for estimation. However, ESMM\textsuperscript{2} does not take into account that Cart does not necessarily occur in the current exposure space, which results in a lack of Cart information related to the conversion in model estimation in the case we discussed above. Thus we have to inject Cart information into the purchase space, as shown by the blue line in Fig~\ref{fig2c}.
    \item \textbf{Right Terms.} The probability space in the estimation one is incorrect. ESMM\textsuperscript{2} takes Cart into consideration over $X_2$, although it took place in the previous exposure space $X_1$, as discussed above. Thus we have to calibrate the sample space, as shown by the orange line in Fig~\ref{fig2c}.
\end{itemize}

The Bad Case for ESMM\textsuperscript{2} is discussed. What happens if the Bad Case doesn't happen for the ground-truth (\textit{i.e.}, Good Case).
\begin{equation}
\begin{split}
    \mathbb{E}_{X_2}[R]&\overset{\textcircled{5}}{=}\mathbb{E}_{X_2}\left[\frac{R}{A}\right]\cdot\mathbb{E}_{X_2}\left[\frac{A}{C}\right]\\
        &=\mathbb{E}_{X_2}[R]\cdot\mathbb{E}_{X_2}\left[\frac{1}{A}\right]\cdot\mathbb{E}_{X_2}[A]\cdot\mathbb{E}_{X_2}\left[\frac{1}{C}\right].
    \end{split}
\end{equation}

\textcircled{5} holds because the user's decision-making path of ``exposure\_click\_Cart\_purchase" is in the same visit in the Good Case which satisfies the assumption of ESMM\textsuperscript{2}. 

Thus, there is no gap in the Good Case.

In summary, there is a significant gap between Cart and other actions (\textit{e.g.} click, purchase). Cart may be related to two sample spaces which leads to the PSC issue. This implies that the in-shop actions does not necessarily satisfy the chain rule, so the strategy of conditional probability cannot be directly employed to manipulate events defined on different sample spaces.

\section{Proposed Method}
In this Section, we propose three approaches to address the PSC issue and improve the performance of Post-Click Conversion Rate estimation.\footnote{The code will be released after publication.}

\subsection{Entire Space Multi-Task Model via Parameter Constraint}

\begin{figure*}
    \centering
    \includegraphics[width=.95\linewidth]{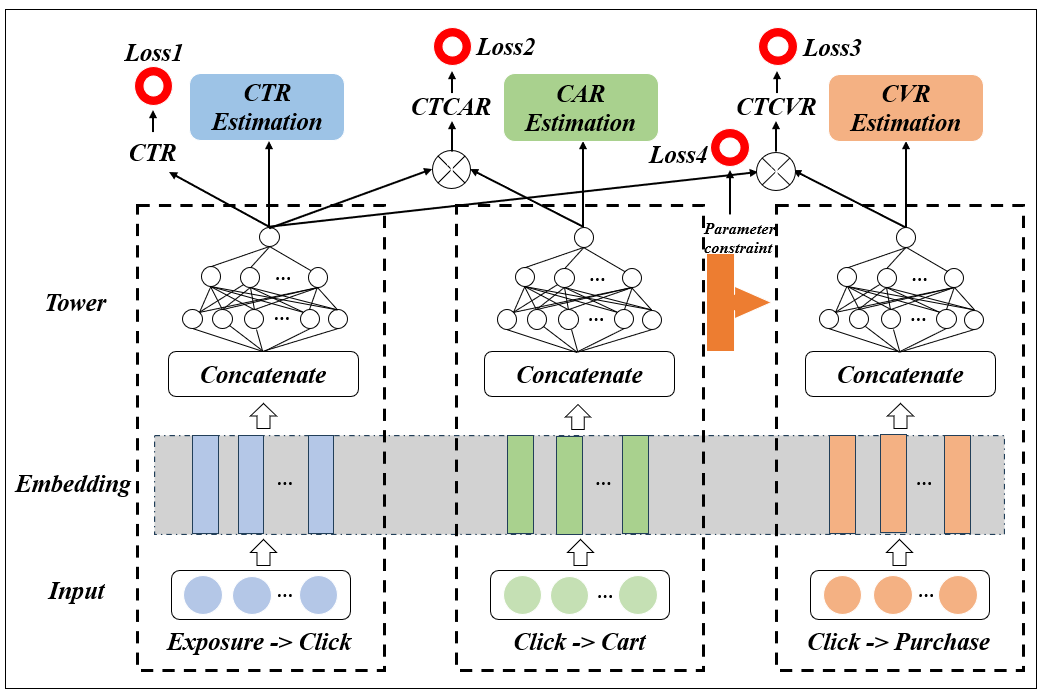}
    \caption{Illustration of ESMC. Loss1 is the CTR loss, loss2 is the CTCAR loss, loss3 is the CTCVR loss, and loss4 is the loss of parameter constraint.}
    \label{fig6}
\end{figure*}

\textbf{Shared Embedding Layer}
First, we build a shared embedding layer to transfer all the sparse ID features and discretized numerical features into dense vectors. The features mainly consist of user features (\textit{e.g.} gender, age, consumption frequency), item features (\textit{e.g.} brand, category, geographic location) and user-item cross features (\textit{e.g.} 
the number of orders in a shop, age-brand). The entire model uses the same embedding, which can be expressed as follows.
\begin{equation}
    \Bar{\mathbf{f}}_i=\mathbf{W}\mathbf{f}_i,
\end{equation}
where $\mathbf{f}_i$ is the i-th one-hot feature and $\mathbf{W}$ denotes the embedding matrix.

\textbf{Constrained Twin Towers} This structure focuses on the decision-making path of ``Cart\_purchase". Since the chain rule of conditional probability cannot describe this path well, we employ a pair of towers to learn the mapping automatically. Specifically, there are one CTCVR tower and one CTCAR (Click-Through Cart Adding Rate) tower. To address the gap in the Left Terms discussed in Section 4, we use the parameter space of CTCAR tower to control that of CTCVR tower. There are three reasons for this.
\begin{itemize}
\item In this way, the information about Cart can be injected into conversion which can couple two information together to fill the gap in the Left Terms, and the proper function can be automatically fitted out by the neural network. 
\item The purchase space is covered by the Cart space, which naturally has a subordinate relationship. 
\item According to our observation, Cart is strongly related to purchase, that is, most of the items in the cart will eventually be bought.
\end{itemize}
Here, we employ KL-divergence to evaluate the distance between two parameter spaces.
\begin{equation}
    D_{KL}(\mathbb{P}(X)\Vert \mathbb{Q}(X))=\mathbb{E}_{X\sim \mathbb{P}(X)}\left[\log \frac{\mathbb{P}(X)}{\mathbb{Q}(X)}\right],
\end{equation}
where $\mathbb{P}(X)$ and $\mathbb{Q}(X)$ express two probability distributions.

\textbf{Sequential Composition Module}
Besides CTCVR tower and CTCAR tower, there is also a CTR tower for CTR estimation. Once we obtain the output of the towers, SCM composes the probability based on the following equation.
\begin{equation}
\begin{split}
    \mathbb{P}(CTCAR)&=\mathbb{P}(CTR)\times\mathbb{P}(CAR),\\
    \mathbb{P}(CTCVR)&=\mathbb{P}(CTR)\times\mathbb{P}(CVR).
\end{split}
\end{equation}

SCM is a nonparameter structure that expresses conditional probability like ESMM.

\textbf{Sample Calibration} 
To calibrate the probability space, we manipulate sample directly. Since some Cart actions have occurred during the user's previous visit, there may be no Cart action prior to the current purchase. We calibrate these samples to correlate them with current purchase behavior according to Session. In this way, the probability space of training samples is explicitly unified that can be represented as the one-to-one mapping function $Q$:
\begin{equation}
    Q: \mathbb{D}^{X_1}\rightarrow\mathbb{D}^{X_2},
\end{equation}
where $\mathbb{D}^{X_1}$ is Cart sample set in the former visit and $\mathbb{D}^{X_2}$ is Cart sample set in the current visit.

After calibration, the ground-truth in the Bad Case becomes:
\begin{equation}
   \mathbb{E}_{X_2}[R]=\mathbb{E}_{X_2}[R|A]\cdot\mathbb{E}_{X_2}[A]\cdot\mathbb{E}_{X_2}\left[\frac{1}{C}\right].
\end{equation}

\begin{table*}[t]
    \centering
    \caption{The basic statistics of datasets. \& denotes the number. M refers to million and K refers to thousand.}
    \begin{tabular}{cccccccc}
    \hline
    Dataset & \&Users & \&Items & \&Clicks & \&Purchases & Total Size & Sparsity of Click & Sparsity of Purchase\\
    \hline
    City 1 & 6M & 110K & 61M & 10M & 1,004M & 0.06096 & 0.01093\\
    City 2 & 3M & 56K & 26M & 4M & 427M & 0.06096 & 0.01108 \\
    City 3 & 4M & 85K & 30M & 5M & 507M & 0.06076 & 0.01067  \\
    City 4 & 3M & 68K & 17M & 2M & 281M & 0.06069 & 0.00908\\
    City 5 & 1M & 30K & 13M & 2M & 216M & 0.06083 & 0.01044\\
    City 6 & 1M & 36K & 11M & 1M & 184M & 0.06103 & 0.01041\\
    \hline
    \end{tabular}
    \label{tab1}
\end{table*}

\textbf{Discussion on ESMC} Finally, the expectation of estimator $\hat{R}$ for ESMC is:
\begin{equation}
\mathbb{E}_{X_2}[\hat{R}]=f_{X_2}(\hat{R},\hat{A})\cdot\mathbb{E}_{X_2}[\hat{A}]\cdot\mathbb{E}_{X_2}\left[\frac{1}{\hat{C}}\right],
\label{eq16}
\end{equation}
where $f_{X_2}$ is an unknown function of $\hat{R}$ and $\hat{A}$ defined on $X_2$. It can be considered as a neural network mapping. Conditional expectation function $\mathbb{E}_{X_2}[R|A]$ can be automatically fitted with neural networks.

The formula is consistent with the ground-truth one and ESMC can perform parameter estimation better in this manner. For the Good Case in Section 4, there is no gap either, which means that ESMC can work well.
Hence, the final training objective to be minimized to obtain parameter set $\Theta$ is as follows.
\begin{equation}
\begin{split}
    L(\Theta)=\omega_1\times L_{CTR}+\omega_2\times L_{CTCVR}+\omega_3\times L_{CTCAR}\\+\omega_4\times D_{KL}(\theta_{CTCAR}\Vert\theta_{CTCVR}),
    \label{eq17}
\end{split}
\end{equation}
where $\omega_1, \omega_2, \omega_3, \omega_4$ are weights of corresponding items, $\theta_{CTCAR}$ and $\theta_{CTCVR}$ are the parameters of CTCAR and CTCVR towers, respectively. Besides, three loss functions $L$ are cross entropy loss shown in \eqref{eq2} with the proper samples of click, Cart and purchase. Fig.~\ref{fig6} shows the structure of ESMC.

\subsection{Entire Space Multi-Task Model with Siamese Network}
 In our practice, we have observed that the conversion rate under the Cart space is very high (more than 80\%). If we adjust the parameter constraints of the twin towers to infinity, it is approximately equivalent to a Siamese Network \cite{chen2022multi} with shared parameters. However, we also find that the change of model performance is not stable along with the increasing of constraint coefficient, which may be due to the constraint conditions affecting the search space of the main task. Therefore, we detach the parameter constraint and directly use the absolute Siamese Network to model CTCVR that is ESMS. Therefore, the constraint in \eqref{eq17} can be removed and the model only focuses on the estimation task. The training objective is expressed as follows.
 \begin{equation}
     L(\Theta)=\omega_1\times L_{CTR}+\omega_2\times L_{CTCVR}+\omega_3\times L_{CTCAR}.
 \end{equation}

\subsection{Entire Space Multi-Task Model in Global Domain}
 We only employ the Cart samples in the recommendation domain in ESMC and ESMS. In fact, items in the shopping cart do not only come from the recommendation domain, but also from the search domain. After a user searches for an item and adds it to the cart, it may still be exposed to the user later by the recommendation system. Therefore, in ESMG, we consider the global domain Cart sample. The structure of model keeps unchanged in ESMG. ESMC\textsuperscript{2} and ESMS\textsuperscript{2} are trained with Cart sample in global domain that are taken into account in later comparative experiments. The training objective is formulated as follows.
\begin{equation}
\begin{split}
    L(\Theta)=\omega_1\times L_{CTR}+\omega_2\times L_{CTCVR}+\omega_3\times L_{CTCAR|rec}\\+\omega_4\times L_{CTCAR|global}+\omega_5\times D_{KL}(\theta_{CTCAR}\Vert\theta_{CTCVR}),
\end{split}
\end{equation}
where $L_{CTCAR|global}$ and $L_{CTCAR|rec}$ means loss functions on Cart over global domain and recommendation domain. Especially, $\omega_5=0$ for ESMS\textsuperscript{2}.

\subsection{Difference among Three Approaches}
\textbf{ESMC \textit{v.s.} ESMS} The difference lies in the way they handle the path of ``Cart\_purchase". In terms of model structure, ESMS is a special case of ESMC. In terms of model performance, ESMS is suitable for scenarios where the Cart space and the purchase space are strongly correlated while ESMC is more suitable for scenarios where the correlation between the Cart space and the purchase space is not as strong. 

In the training stage, it takes a lot of time to adjust the constraint coefficient. Besides, due to the dependence between the twin towers, it is difficult to train the two towers in parallel, which increases the overhead. In the inference phase, as the twin networks of ESMS share parameters, only parameters in one tower needs to be stored, which significantly reduces the number of parameters and memory occupied by the model. This makes deployment of the model to online platforms more efficient.

\textbf{ESMC\&ESMS \textit{v.s.} ESMG (ESMC\textsuperscript{2}\&ESMS\textsuperscript{2})} The difference is whether the Cart sample comes from the global domain or the recommendation domain. Only using sample from the recommendation domain absolutely conforms to the basic assumptions of proposed methodology in Section 4. However, the use of global Cart samples may relax the basic assumptions that affect the performance of the model. But at the same time, considering Cart samples from the global domain can supplement the information that helps improve the generalization of algorithm. The degree of information gain and constraint relaxation is related to the correlation between the search domain and the recommendation domain.

In conclusion, we propose three approaches (four models) to address the PSC issue. Considering the online performance and the cost of model deployment, we finally choose ESMS to deploy on the online environment. Later, our proposed models are collectively referred to as ESMC-family.

\begin{table*}[htbp]
  \centering
  \caption{Comparison with SOTA multi-task learning baselines. The best results are shown in \textbf{Bold} and the second best results are shown in \textit{Italic}. Improvement is calculated as the relative increase of our best result compared to the best result in the baselines.}
    \begin{tabular}{cccccccccc}
    \hline
          & \multicolumn{3}{c}{City 1} & \multicolumn{3}{c}{City 2} & \multicolumn{3}{c}{City 3} \\
          \hline
          & CTR-AUC   & CTCVR-AUC & CVR-AUC   & CTR-AUC   & CTCVR-AUC & CVR-AUC   & CTR-AUC   & CTCVR-AUC & CVR-AUC \\
          \hline
    Shared Bottom & 0.73025  & 0.82245  & 0.71242  & 0.72674  & 0.81406  & 0.70152  & 0.73103  & 0.82364  & 0.70824  \\
    ESMM  & 0.72894  & 0.82395  & 0.71949  & 0.72546  & 0.81649  & 0.71002  & 0.73010  & 0.82532  & 0.71606  \\
    MMOE  & 0.72963  & 0.82254  & 0.71327  & 0.72592  & 0.81450  & 0.70239  & 0.73065  & 0.82366  & 0.70894  \\
    ESMM\textsuperscript{2} & 0.72995  & 0.83306  & 0.76824  & 0.72681  & 0.82692  & 0.75955  & 0.73100  & 0.83343  & 0.76117  \\
    \hline
    ESMS  & \textit{0.73093 } & \textit{0.83563 } & \textbf{0.77048 } & \textit{0.72697 } & \textit{0.82849 } & \textbf{0.76069 } & \textit{0.73140 } & \textit{0.83599 } & \textbf{0.76385 } \\
    ESMC  & \textbf{0.73111 } & \textbf{0.83594 } & \textit{0.76862 } & \textbf{0.72774 } & \textbf{0.82936 } & \textit{0.75987 } & \textbf{0.73176 } & \textbf{0.83637 } & \textit{0.76339 } \\
    \hline
    Improvement & 0.118\% & 0.346\% & 0.292\% & 0.128\% & 0.295\% & 0.150\% & 0.100\% & 0.353\% & 0.352\% \\
    \hline
    \hline
          & \multicolumn{3}{c}{City 4} & \multicolumn{3}{c}{City 5} & \multicolumn{3}{c}{City 6} \\
          \hline
          & CTR-AUC   & CTCVR-AUC & CVR-AUC  & CTR-AUC   & CTCVR-AUC & CVR-AUC   & CTR-AUC   & CTCVR-AUC & CVR-AUC \\
          \hline
    Shared Bottom & 0.72465  & 0.81497  & 0.69693  & 0.72902  & 0.82326  & 0.71015  & 0.73349  & 0.83099  & 0.71435  \\
    ESMM  & 0.72367  & 0.81704  & 0.70419  & 0.72777  & 0.82608  & 0.71966  & 0.73262  & 0.83156  & 0.72050  \\
    MMOE  & 0.72419  & 0.81481  & 0.69664  & 0.72836  & 0.82460  & 0.71322  & 0.73349  & 0.83076  & 0.71518  \\
    ESMM\textsuperscript{2} & \textit{0.72515 } & 0.82627  & 0.74893  & 0.72866  & 0.83532  & 0.76531  & 0.73336  & 0.84168  & 0.76769  \\
    \hline
    ESMS  & 0.72484  & \textit{0.82782 } & \textit{0.75016 } & \textit{0.72959 } & \textit{0.83716 } & \textit{0.76552 } & \textit{0.73433 } & \textit{0.84305 } & \textit{0.76847 } \\
    ESMC  & \textbf{0.72806 } & \textbf{0.84098 } & \textbf{0.75788 } & \textbf{0.73009 } & \textbf{0.83737 } & \textbf{0.76762 } & \textbf{0.73493 } & \textbf{0.84334 } & \textbf{0.76871 } \\
    \hline
    Improvement & 0.401\% & 1.780\% & 1.195\% & 0.147\% & 0.245\% & 0.302\% & 0.196\% & 0.197\% & 0.133\% \\
    \hline
    \end{tabular}%
  \label{tab2}%
\end{table*}%

\begin{table}[t]
    \centering
    \caption{Results of comparison of ESMS\&ESMC and ESMS\textsuperscript{2}\&ESMC\textsuperscript{2}. Improvement is calculated as the relative increase of models in global domain compared to models in recommender domain.}
    \begin{tabular}{cccc}
    \hline
     \multicolumn{4}{c}{City 1} \\
     \hline
     & CTR-AUC & CTCVR-AUC & CVR-AUC \\
     \hline
     ESMS\textsuperscript{2} & 0.73116 & 0.83589 & 0.77066  \\
     Improvement & 0.031\% & 0.031\% & 0.023\%   \\
     ESMC\textsuperscript{2} &  0.74015 & 0.84166 & 0.76966  \\
     Improvement & 1.236\% & 0.684\% & 0.135\%  \\
    \hline
    \hline
     \multicolumn{4}{c}{City 2} \\
     \hline
     & CTR-AUC & CTCVR-AUC & CVR-AUC \\
     \hline
     ESMS\textsuperscript{2} & 0.72746 & 0.82933 & 0.76082 \\
     Improvement & 0.067\% & 0.101\% & 0.017\%   \\
     ESMC\textsuperscript{2} &  0.73518  & 0.83615 & 0.76039  \\
     Improvement & 1.022\% & 0.818\% & 0.684\%  \\
    \hline
    \hline
     \multicolumn{4}{c}{City 3} \\
     \hline
     & CTR-AUC & CTCVR-AUC & CVR-AUC \\
     \hline
     ESMS\textsuperscript{2} & 0.73166 & 0.83621 & 0.76388  \\
     Improvement & 0.035\% & 0.026\% & 0.004\%  \\
     ESMC\textsuperscript{2} & 0.73835 & 0.84272 & 0.76496   \\
     Improvement &  0.900\% & 0.759\% & 0.205\% \\
    \hline
    \end{tabular}
    \label{tab4}
\end{table}

\section{Experiments}
We conduct extensive experiments to evaluate the performance of ESMC-family and the following research questions (RQs) are answered:
\begin{itemize}
    \item \textbf{RQ1} Do ESMC and ESMS outperform state-of-the-art multi-task estimators?
    \item \textbf{RQ2} What is the difference between the performance of ESMC and that of ESMS? How to choose the proper one?
    \item \textbf{RQ3} What is the difference between the performance of ESMC\&ESMS and that of ESMG?
    \item \textbf{RQ4} How do critical components affect the performance of ESMC-family?
    \item \textbf{RQ5} Can ESMC-family address PSC issue?
    \item \textbf{RQ6} Does ESMC-family work in real large-scale online recommendation scenarios?
\end{itemize}
    
\subsection{Experimental Settings}
\textbf{Datasets\footnote{To the best of our knowledge, there are no public datasets suited for this task, and we believe that a huge amount of data can significantly verify the performance of recommender. We will release our datasets for future research after desensitization and checking.}} We collect six offline datasets by collecting the users' feedback logs from six different cities between April 21, 2023 and May 10, 2023 from a large-scale online platform named Eleme, Alibaba's takeaway platform that produces nearly one billion behavioral data per day. The statistics of the offline datasets are listed in Table~\ref{tab1}.

\textbf{Metrics} To evaluate the performance of our proposed method, we select three widely used metrics for offline test, \textit{i.e.}, CTR-AUC, CTCVR-AUC and CVR-AUC for CTR, CTCVR and CVR estimation. 
\begin{equation}
    \text{AUC}=\frac{1}{|P||N|}\Sigma_{p\in P}\Sigma_{n\in N}I(\Theta(p)>\Theta(n)),
\end{equation}
where $P$ and $N$ denote positive sample set and negative sample set, respectively. $\Theta$ is the estimator function and $I$ is the indicator function.

\textbf{Baselines}
The representative state-of-the-art approaches are listed as follows. All models are equipped with DIN \cite{zhou2018deep} and ETA \cite{chen2021end}, sequential recommenders which propose interest extraction from users' historical behaviors with attention mechanism \cite{vaswani2017attention}, to extract user's long/short-term interests.
\begin{itemize}
    \item \textbf{Shared Bottom} To tackle multi-task learning, the output layer is replaced by two fully-connected towers for the corresponding two tasks. Moreover, the two tasks share the same bottom to learn common features \cite{tang2020progressive}.
    \item \textbf{ESMM} ESMM models conversion rate with conditional probability on the user decision-making path of “exposure\_click\_purchase” without considering the in-shop actions.
    \item \textbf{MMoE} MMoE employs several expert networks with gate controllers to leverage different downstream tasks \cite{ma2018modeling}.
    \item \textbf{ESMM\textsuperscript{2}} ESMM\textsuperscript{2} models conversion rate with conditional probability on the user decision-making graph of “exposure\_click\_in-shop action\_purchase” that is degraded by PSC issue.
\end{itemize}

\begin{figure*}[htbp]
	\centering
	\subfloat[CTR-AUC on City 6 dataset.]{\includegraphics[width=.3\linewidth]{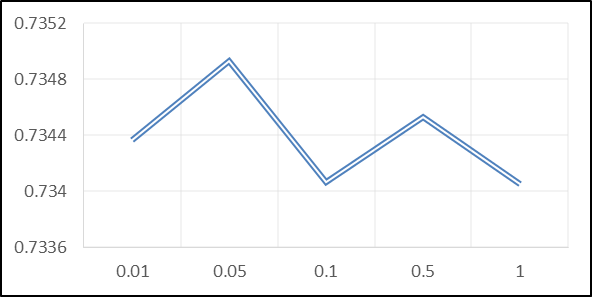}}\hspace{5pt}
	\subfloat[CTCVR-AUC on City 6 dataset.]{\includegraphics[width=.3\linewidth]
 {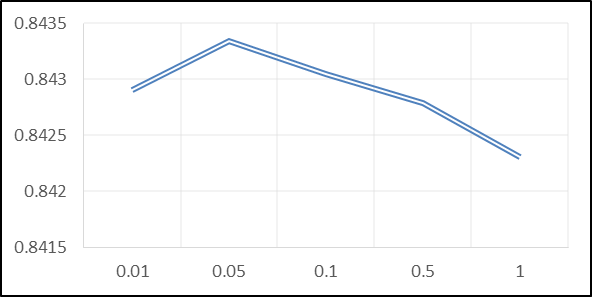}}\hspace{5pt}
 \subfloat[CVR-AUC on City 6 dataset.]{\includegraphics[width=.3\linewidth]
 {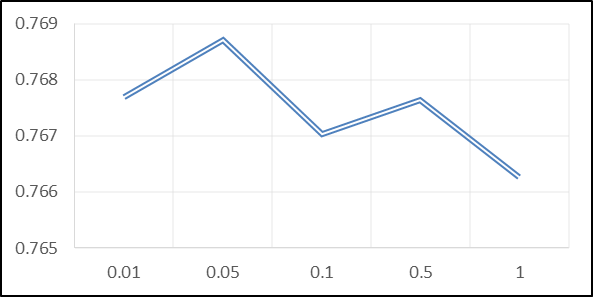}}
 \\
	\subfloat[CTR-AUC on City 4 dataset.]{\includegraphics[width=.3\linewidth]{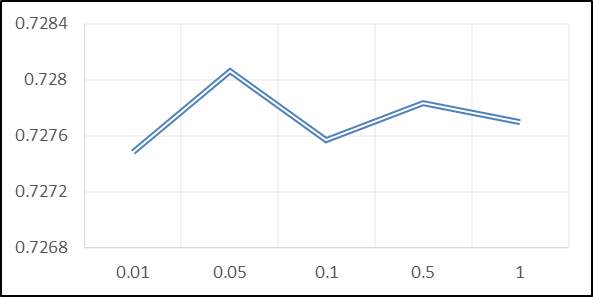}}\hspace{5pt}
	\subfloat[CTCVR-AUC on City 4 dataset.]{\includegraphics[width=.3\linewidth]
 {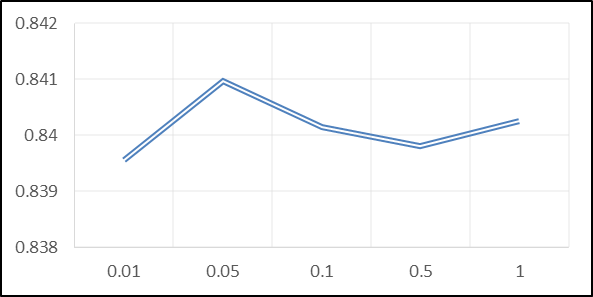}}\hspace{5pt}
 \subfloat[CVR-AUC on City 4 dataset.]{\includegraphics[width=.3\linewidth]
 {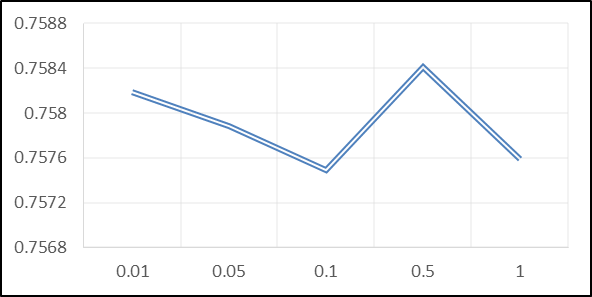}}
	\caption{Sensitivity of coefficient of parameter constraint in ESMC. The horizontal axis represents the parameter, and the vertical axis represents AUC.}
        \label{fig8}
\end{figure*}

\textbf{Training Protocol} All models in this paper are implemented with Tensorflow 1.12 in Python 2.7 environment. All models are trained in a chief-worker distributed structure with 1600 CPUs \cite{gu2019tiresias}. AdagradDecay \cite{antonakopoulos2022adagrad} is chosen as our optimizer for model training and activation function is set to LeakyReLU \cite{frei2022implicit}. The initial learning rate is set to 0.005, batch size is set to 1024, and training epoch is set to 1 because of one-epoch phenomenon \cite{zhang2022towards}. The top-k for behavior sequence in ETA is set to 50. All models are employed the same warm-up technique to maintain the training stability. In training stage, the weights of CTR, CVR and CAR loss are equal to 1. The coefficient of parameter constraint for ESMC is selected from \{0.01, 0.05, 0.1, 0.5, 1.0\} and the coefficient of CAR loss in global domain is selected from \{0.1, 0.3, 0.5, 0.7, 1.0\}.

\subsection{RQ1\&RQ2: Comparison with Baselines}
Table~\ref{tab2} indicates that ESMS and ESMC outperform all SOTA baselines in terms of CTR-AUC, CTCVR-AUC and CVR-AUC on six real-world datasets. All the best results come from our methods and all the second best results come from our methods except comparison on City 4 dataset in terms of CTR-AUC. Besides, the Improvement proves that our proposed approaches achieve a further significant improvement\footnote{Note that the 0.1\% AUC gain is already considerable in large-scale industrial recommender \cite{zhou2018deep, du2022basm}.}. Especially, on City 4 dataset, ESMC achieves a huge improvement compared to the best baseline ESMM\textsuperscript{2}, \textit{i.e.}, CTR-AUC +0.401\%, CTCVR-AUC +1.780\%, and CVR-AUC +1.195\%. 

Considering the baselines, Shared Bottom performs well on the CTR estimation task. Because of the low coupling of the parameters of CTR and CVR estimation in Shared Bottom, the prediction of the two tasks is more independent \cite{chen2023cross}. At the same time, the number of click samples is more than that of conversion samples, so it can achieve relatively high accuracy in the CTR task. However, this does not mean that other baselines are poor, quite the contrary, MMoE and ESMM have a wide range of applications in industry \cite{gong2022real, wang2020next}. On average, Shared Bottom's CVR prediction is the worst, and in industrial practice, Conversion Rate is generally more important than Click-Through Rate. In our proposed approach, CTR and CVR tasks are linked through their intrinsic relationships and in-shop information, while maintaining a high degree of independence between CTR and CVR tasks, thus achieving better results on CTR prediction. 

For CTCVR-AUC and CVR-AUC, ESMM\textsuperscript{2} gives the best results among baselines. Especially, Post-Click Conversion Rate has improved significantly, with an average increase of more than 5\%! This fully illustrates the importance of introducing Cart behavior to CVR estimation, and also shows that our work on enhancing ESMM\textsuperscript{2} is critical and meaningful for recommenders. In our proposed methods, by solving ESMM\textsuperscript{2}'s bad case: Probability Space Confusion issue, we further improve the performance of CVR estimation of ESMM\textsuperscript{2}, and achieve SOTA in all experiments!

Considering ESMS \textit{v.s.} ESMC, ESMC is better than ESMS on average. Sometimes, ESMS performs better in CVR prediction task, which may be related to the final Conversion Rate of goods in shopping carts for users in different cities. ESMC can adjust the coefficient of parameter constraint to manipulate the model performance. Therefore, if you are looking for better results, ESMC is recommended. If you don't want to adjust hyper-parameters, ESMS is recommended. To facilitate version iteration, we chose to deploy ESMS to serve users.

\begin{figure*}[htbp]
	\centering
	\subfloat[CTR-AUC on City 1 dataset.]{\includegraphics[width=.3\linewidth]{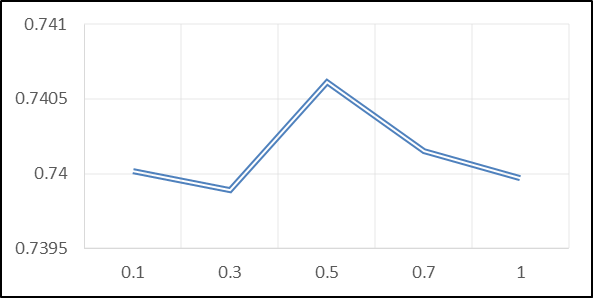}}\hspace{5pt}
	\subfloat[CTCVR-AUC on City 1 dataset.]{\includegraphics[width=.3\linewidth]
 {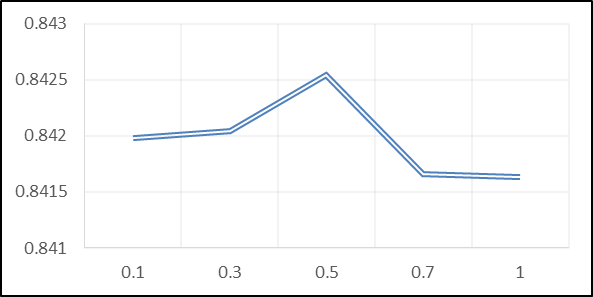}}\hspace{5pt}
 \subfloat[CVR-AUC on City 1 dataset.]{\includegraphics[width=.3\linewidth]
 {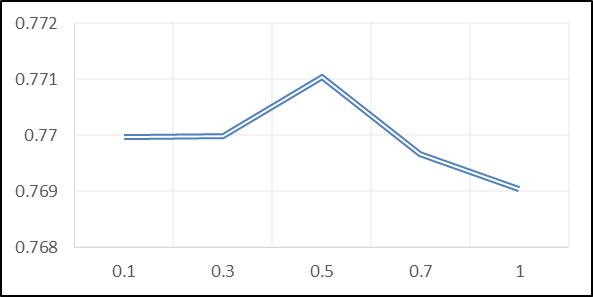}}
 \\
	\subfloat[CTR-AUC on City 2 dataset.]{\includegraphics[width=.3\linewidth]{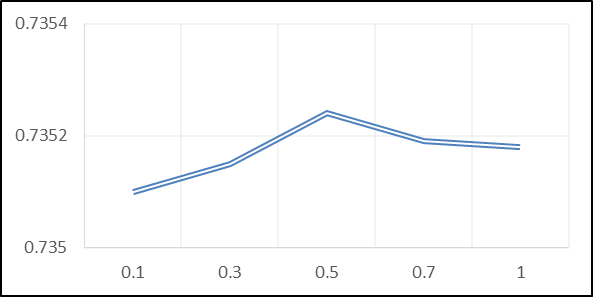}}\hspace{5pt}
	\subfloat[CTCVR-AUC on City 2 dataset.]{\includegraphics[width=.3\linewidth]
 {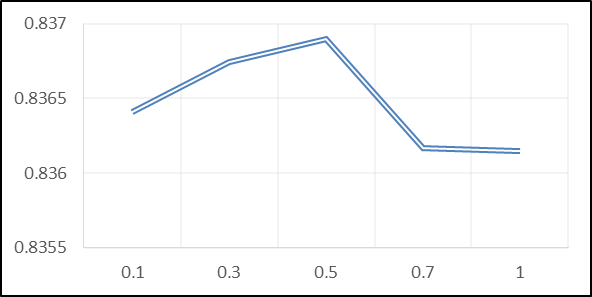}}\hspace{5pt}
 \subfloat[CVR-AUC on City 2 dataset.]{\includegraphics[width=.3\linewidth]
 {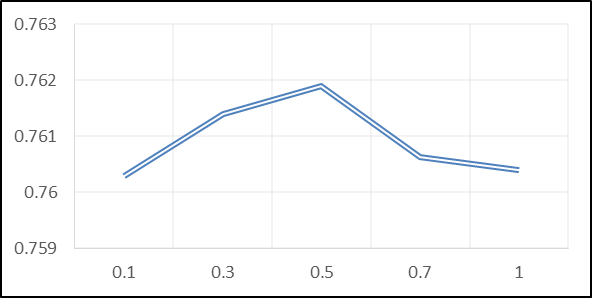}}
	\caption{Sensitivity of weight of global domain loss in ESMC\textsuperscript{2}. The horizontal axis represents the weight, and the vertical axis represents AUC.}
        \label{fig9}
\end{figure*}

\subsection{RQ3: Recommender Domain \textit{v.s.} Global Domain}
Table~\ref{tab4} indicates that ESMG absolutely outperforms ESMS\&ESMC in our experiments. However, the Improvement of ESMS\textsuperscript{2} is quite small. On the contrary, ESMC\textsuperscript{2} achieve a considerable improvement compared with ESMC. Especially, CTR-AUC raises over 1\% and CTCVR-AUC raises over 0.7\% on average! The performance of ESMS\textsuperscript{2} is not significant probably because CTCVR and CTCAR are trainined with the shared parameters. After bringing global domain Cart samples, the sample distribution becomes more complex, which is difficult to fit. However, ESMC can handle this problem. As a result, ESMC\textsuperscript{2} has made a huge improvement. Thus, on the leaderboard of ESMC-family, ESMC\textsuperscript{2} is the best while ESMS is the worst.

\begin{figure}[htbp]
	\centering
	\subfloat[CTR-AUC on City 3 dataset.]{\includegraphics[width=.31\columnwidth]{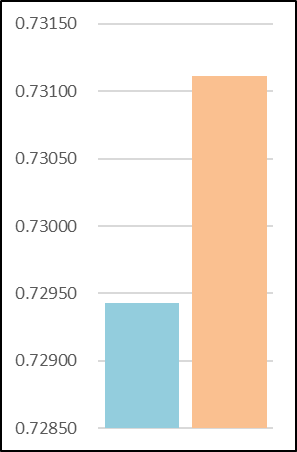}}\hspace{5pt}
	\subfloat[CTCVR-AUC on City 3 dataset.]{\includegraphics[width=.31\columnwidth]
 {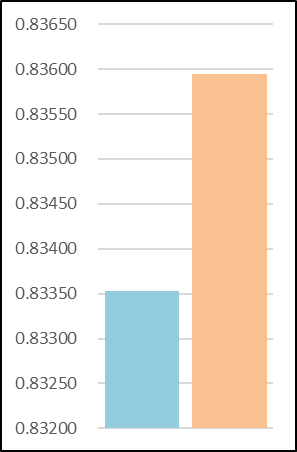}}\hspace{5pt}
 \subfloat[CVR-AUC on City 3 dataset.]{\includegraphics[width=.31\columnwidth]
 {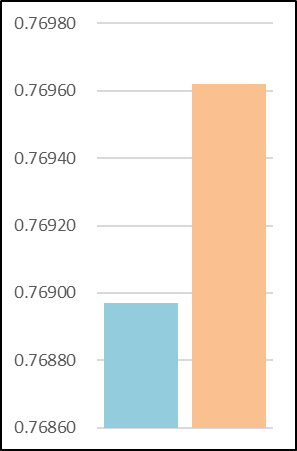}}
 \\
	\subfloat[CTR-AUC on City 4 dataset.]{\includegraphics[width=.31\columnwidth]{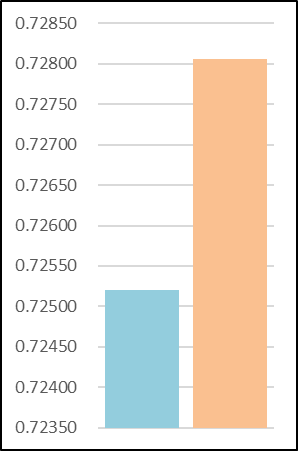}}\hspace{5pt}
	\subfloat[CTCVR-AUC on City 4 dataset.]{\includegraphics[width=.31\columnwidth]
 {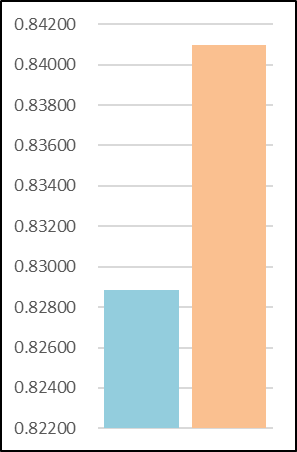}}\hspace{5pt}
 \subfloat[CVR-AUC on City 4 dataset.]{\includegraphics[width=.31\columnwidth]
 {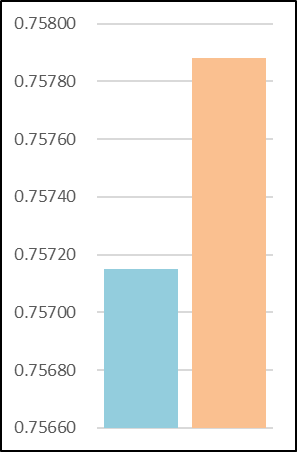}}
	\caption{Ablation study on Sample Calibration in ESMC. The blue bar is for ESMC- while the orange bar is for ESMC.}
        \label{fig10}
\end{figure}

\begin{table*}[t]
    \centering
    \caption{The comparison results on the Bad Case and the Good Case in terms of CVR-AUC. Improvement is calculated as the relative
increase of ESMS compared to ESMM\textsuperscript{2}.}
    \begin{tabular}{ccccccc}
    \hline
    \multicolumn{7}{c}{Bad Case} \\
    \hline
         & City 1 & City 2 & City 3 & City 4 & City 5 & City 6 \\
    \hline
      ESMM\textsuperscript{2} & 0.65848 & 0.66022 & 0.66176 & 0.67414 & 0.66887 & 0.67514 \\
      ESMS & 0.70224 & 0.70220 & 0.70426 & 0.71108 & 0.71114 & 0.71213  \\
      Improvement & 6.645\% & 6.358\% & 6.422\% & 5.479\% & 6.319\% & 5.478\% \\
      \hline
    \hline
    \multicolumn{7}{c}{Good Case} \\
    \hline
         & City 1 & City 2 & City 3 & City 4 & City 5 & City 6 \\
    \hline
      ESMM\textsuperscript{2} & 0.68472 & 0.69251 & 0.68935 & 0.72147 & 0.69270 & 0.70194 \\
      ESMS & 0.72341 & 0.72469 & 0.72448 & 0.74979 & 0.72545 & 0.72902 \\
      Improvement & 5.650\% & 4.646\% & 5.096\% & 3.925\% & 4.727\% & 3.857\% \\
      \hline
    \end{tabular}
    \label{tab5}
\end{table*}

\begin{table*}[t]
    \centering
    \caption{Online A/B performances for consecutive 7 days on Eleme.}
    \begin{tabular}{ccccccccc}
    \hline
      Day& 1&2&3&4&5&6&7&Avg. \\ 
    \hline
       NU & +0.75\% & +1.05\% & +0.45\% & +1.14\% & +0.40\% & +1.06\% & +0.33\% & +0.74\% \\
       NO & +1.42\% & +1.41\% & +0.67\% & +1.41\% & +0.54\% & +1.38\% & +0.56\% & +1.05\% \\
       OR & +1.57\% & +1.45\% & +0.76\% & +1.44\% & +0.59\% & +1.33\% & +0.75\% & +1.12\%  \\
       NG & +1.33\% & +0.88\% & +0.39\% & +1.77\% & +0.62\% & +1.44\% & +0.63\% & +1.00\% \\
    \hline
    \end{tabular}
    \label{tab3}
\end{table*}

\subsection{RQ4: Parameter Sensitivity and Ablation Study}
Fig.~\ref{fig8} illustrates the curves of model performance varying with coefficient of parameter constraint in ESMC. It can be seen that the performance of model shows a fluctuating curve, and there is no obvious increasing/decreasing tendency. Especially, ESMC generally performs well when the coefficient of parameter constraint being set to 0.05 in our experiments. However, this does not mean that 0.05 is the best parameter for ESMC to trace the underlying relationship in path of ``exposure\_click\_Cart\_purchase". Because the optimization objective of ESMC is a linear combination of multiple sub-objectives. Optimizing multiple sub-objectives at the same time will affect the training of the model, and it is difficult to find the Pareto Optimal \cite{deb2005searching}. Therefore, we need to adjust this parameter carefully when using ESMC. Fortunately, ESMC may offer better returns. 

Fig.~\ref{fig9} shows the curves of model performance varying with weight of all domain loss in ESMC\textsuperscript{2}. It can be seen that the trend of curve is to rise first followed by a decreasing. Especially,  ESMC\textsuperscript{2} generally performs well when the weight of global domain loss being set to 0.5 in our experiments. This result is in line with the distribution of traffic in our business. Therefore, we recommend to set the weight of the loss function directly according to the traffic distribution of different domains.

Fig.~\ref{fig10} presents the ablation study on Sample Calibration. We select ESMC as the research object because it is the initial model in ESMC-family. ESMC detached Sample Calibration is named ESMC-. The results show that Sample Calibration can improve the performance significantly. We observe that when it is removed, CTCVR-AUC would fall by 0.11\% on average, which implies that Sample Calibration is a simple and efficient strategy to maintain the consistency of sample selection and probability space.

\subsection{RQ5: Case Study.}
In Section 4, we have discussed the Bad Case and the Good Case of ESMM\textsuperscript{2}. ESMC-family is tailored for PSC issue to handle the Bad Case. Here, we do experiments to prove that the model does solve PSC issue and also improves the performance on the Good Case. 
We select conversion samples (purchase label = 1) to be divided into two groups: 1) Bad Case: Cart and purchase are not in the same exposure space and 2) Good Case: Cart and purchase are in the same exposure space. Because all samples' conversion labels are 1, which means that all samples' click labels are also 1, considering CTR-AUC  doesn't make sense. Besides, CTCVR-AUC is equal to CVR-AUC here. Therefore, we just consider CVR-AUC in this experiment. Table~\ref{tab5} shows that ESMS absolutely outperforms ESMM\textsuperscript{2} on both the Good Case and the Bad Case. There is no doubt that ESMC-family can address PSC issue perfectly.

Especially, on average, ESMS improves CVR-AUC by over 6\% on the Bad Case and near 5\% on the Good Case!!! The reason why the model can achieve a huge improvement on the Good Case may be that the parameter constraint strategy introduces more information about Cart, which is helpful for the fitting of the high-dimensional function of $\hat{a}$ in \eqref{eq16}.

\subsection{RQ6: Online A/B Test}
From June 22, 2023 to June 28, 2023, we conducted a seven-day online experiment by deploying ESMS (ESMS in recommender domain) to the recommendation scenario on the homepage of Eleme. The online base model is a variation of MMoE with long/short-term behavior sequence module like DIN and ETA. Here, we select four business-related metrics: Number of Paying Users (NU), Number of Orders (NO), Order Rate (OR) and Net GMV (NG, a measure of net profit) to evaluate the performance in online environment. Results of strictly online A/B tests are shown in Table~\ref{tab3}. We can see that the proposed ESMS, an alternative of ESMC, consistently outperforms the base model. On average, our method improves NU by 0.74\%, NO by 1.05\%, OR by 1.12\% and NG by 1.00\% compared with the base model, which demonstrates the effectiveness of ESMC-families in large-scale online recommendation system and the proposed approach has been deployed on the homepage of Eleme, Alibaba's online takeaway platform serving more than one billion recommendation requests per day.

\section{Conclusion and Future Work}
In this paper, we report Probability Space Confusion issue in the chain rule of conditional probability about in-shop actions (\textit{e.g.} Cart/Wish-list) and present a mathematical explanation to illustrate the gap between ground-truth and estimation. We show that the key points are inconsistent sample space and decoupling of purchase and Cart. Based on this, we propose a novel Entire Space Multi-Task Model via Parameter Constraint and two alternatives: Entire Space Multi-Task Model with Siamese Network and Entire Space Multi-Task Model in Global Domain (ESMC\textsuperscript{2} and ESMS\textsuperscript{2}) to address PSC problem. Extensive offline experiments prove the performance of our proposed approaches and the seven-day online A/B test shows that ESMS yields significant financial benefits to our business. To support the future research, we discuss the advantages and disadvantages of three proposed approaches and decide to release real-world datasets and code. The future work may include exploring a better parameter constraint strategy to make ESMC's training more stable and a better method to calibrate sample space.

\bibliographystyle{IEEEtranS.bst}
\bibliography{ref}

% Generated by IEEEtranS.bst, version: 1.12 (2007/01/11)
\begin{thebibliography}{10}
\providecommand{\url}[1]{#1}
\csname url@samestyle\endcsname
\providecommand{\newblock}{\relax}
\providecommand{\bibinfo}[2]{#2}
\providecommand{\BIBentrySTDinterwordspacing}{\spaceskip=0pt\relax}
\providecommand{\BIBentryALTinterwordstretchfactor}{4}
\providecommand{\BIBentryALTinterwordspacing}{\spaceskip=\fontdimen2\font plus
\BIBentryALTinterwordstretchfactor\fontdimen3\font minus
  \fontdimen4\font\relax}
\providecommand{\BIBforeignlanguage}[2]{{%
\expandafter\ifx\csname l@#1\endcsname\relax
\typeout{** WARNING: IEEEtranS.bst: No hyphenation pattern has been}%
\typeout{** loaded for the language `#1'. Using the pattern for}%
\typeout{** the default language instead.}%
\else
\language=\csname l@#1\endcsname
\fi
#2}}
\providecommand{\BIBdecl}{\relax}
\BIBdecl

\bibitem{antonakopoulos2022adagrad}
K.~Antonakopoulos, P.~Mertikopoulos, G.~Piliouras, and X.~Wang, ``Adagrad
  avoids saddle points,'' in \emph{International Conference on Machine
  Learning}.\hskip 1em plus 0.5em minus 0.4em\relax PMLR, 2022, pp. 731--771.

\bibitem{bansal2016ask}
T.~Bansal, D.~Belanger, and A.~McCallum, ``Ask the gru: Multi-task learning for
  deep text recommendations,'' in \emph{proceedings of the 10th ACM Conference
  on Recommender Systems}, 2016, pp. 107--114.

\bibitem{chen2021attentive}
L.~Chen, J.~Cao, H.~Chen, W.~Liang, H.~Tao, and G.~Zhu, ``Attentive multi-task
  learning for group itinerary recommendation,'' \emph{Knowledge and
  Information Systems}, vol.~63, no.~7, pp. 1687--1716, 2021.

\bibitem{chen2022multi}
L.~Chen, Z.~Li, T.~Xu, H.~Wu, Z.~Wang, N.~J. Yuan, and E.~Chen, ``Multi-modal
  siamese network for entity alignment,'' in \emph{Proceedings of the 28th ACM
  SIGKDD conference on knowledge discovery and data mining}, 2022, pp.
  118--126.

\bibitem{chen2021end}
Q.~Chen, C.~Pei, S.~Lv, C.~Li, J.~Ge, and W.~Ou, ``End-to-end user behavior
  retrieval in click-through rateprediction model,'' \emph{arXiv preprint
  arXiv:2108.04468}, 2021.

\bibitem{chen2023cross}
X.~Chen, Z.~Cheng, S.~Xiao, X.~Zeng, and W.~Huang, ``Cross-domain augmentation
  networks for click-through rate prediction,'' \emph{arXiv preprint
  arXiv:2305.03953}, 2023.

\bibitem{deb2005searching}
K.~Deb and H.~Gupta, ``Searching for robust pareto-optimal solutions in
  multi-objective optimization,'' in \emph{International conference on
  evolutionary multi-criterion optimization}.\hskip 1em plus 0.5em minus
  0.4em\relax Springer, 2005, pp. 150--164.

\bibitem{du2022basm}
B.~Du, S.~Lin, J.~Gao, X.~Ji, M.~Wang, T.~Zhou, H.~He, J.~Jia, and N.~Hu,
  ``Basm: A bottom-up adaptive spatiotemporal model for online food ordering
  service,'' \emph{arXiv preprint arXiv:2211.12033}, 2022.

\bibitem{feng2022social}
X.~Feng, Z.~Liu, W.~Wu, and W.~Zuo, ``Social recommendation via deep neural
  network-based multi-task learning,'' \emph{Expert Systems with Applications},
  vol. 206, p. 117755, 2022.

\bibitem{frei2022implicit}
S.~Frei, G.~Vardi, P.~L. Bartlett, N.~Srebro, and W.~Hu, ``Implicit bias in
  leaky relu networks trained on high-dimensional data,'' \emph{arXiv preprint
  arXiv:2210.07082}, 2022.

\bibitem{gao2019neural}
C.~Gao, X.~He, D.~Gan, X.~Chen, F.~Feng, Y.~Li, T.-S. Chua, and D.~Jin,
  ``Neural multi-task recommendation from multi-behavior data,'' in \emph{2019
  IEEE 35th international conference on data engineering (ICDE)}.\hskip 1em
  plus 0.5em minus 0.4em\relax IEEE, 2019, pp. 1554--1557.

\bibitem{gao2023rec4ad}
J.~Gao, S.~Han, H.~Zhu, S.~Yang, Y.~Jiang, J.~Xu, and B.~Zheng, ``Rec4ad: A
  free lunch to mitigate sample selection bias for ads ctr prediction in
  taobao,'' \emph{arXiv preprint arXiv:2306.03527}, 2023.

\bibitem{gong2022real}
X.~Gong, Q.~Feng, Y.~Zhang, J.~Qin, W.~Ding, B.~Li, P.~Jiang, and K.~Gai,
  ``Real-time short video recommendation on mobile devices,'' in
  \emph{Proceedings of the 31st ACM International Conference on Information \&
  Knowledge Management}, 2022, pp. 3103--3112.

\bibitem{gu2019tiresias}
J.~Gu, M.~Chowdhury, K.~G. Shin, Y.~Zhu, M.~Jeon, J.~Qian, H.~Liu, and C.~Guo,
  ``Tiresias: A $\{$GPU$\}$ cluster manager for distributed deep learning,'' in
  \emph{16th USENIX Symposium on Networked Systems Design and Implementation
  (NSDI 19)}, 2019, pp. 485--500.

\bibitem{gu2021estimating}
T.~Gu, K.~Kuang, H.~Zhu, J.~Li, Z.~Dong, W.~Hu, Z.~Li, X.~He, and Y.~Liu,
  ``Estimating true post-click conversion via group-stratified counterfactual
  inference,'' 2021.

\bibitem{gupte2020automated}
R.~Gupte, S.~Rege, S.~Hawa, Y.~Rao, and R.~Sawant, ``Automated shopping cart
  using rfid with a collaborative clustering driven recommendation system,'' in
  \emph{2020 Second International Conference on Inventive Research in Computing
  Applications (ICIRCA)}.\hskip 1em plus 0.5em minus 0.4em\relax IEEE, 2020,
  pp. 400--404.

\bibitem{hadash2018rank}
G.~Hadash, O.~S. Shalom, and R.~Osadchy, ``Rank and rate: multi-task learning
  for recommender systems,'' in \emph{Proceedings of the 12th ACM Conference on
  Recommender Systems}, 2018, pp. 451--454.

\bibitem{kumar2015predicting}
R.~Kumar, S.~M. Naik, V.~D. Naik, S.~Shiralli, V.~Sunil, and M.~Husain,
  ``Predicting clicks: Ctr estimation of advertisements using logistic
  regression classifier,'' in \emph{2015 IEEE international advance computing
  conference (IACC)}.\hskip 1em plus 0.5em minus 0.4em\relax IEEE, 2015, pp.
  1134--1138.

\bibitem{lee2012estimating}
K.-c. Lee, B.~Orten, A.~Dasdan, and W.~Li, ``Estimating conversion rate in
  display advertising from past erformance data,'' in \emph{Proceedings of the
  18th ACM SIGKDD international conference on Knowledge discovery and data
  mining}, 2012, pp. 768--776.

\bibitem{li2021attentive}
D.~Li, B.~Hu, Q.~Chen, X.~Wang, Q.~Qi, L.~Wang, and H.~Liu, ``Attentive capsule
  network for click-through rate and conversion rate prediction in online
  advertising,'' \emph{Knowledge-Based Systems}, vol. 211, p. 106522, 2021.

\bibitem{li2020multi}
H.~Li, Y.~Wang, Z.~Lyu, and J.~Shi, ``Multi-task learning for recommendation
  over heterogeneous information network,'' \emph{IEEE Transactions on
  Knowledge and Data Engineering}, vol.~34, no.~2, pp. 789--802, 2020.

\bibitem{lin2022feature}
Z.~Lin, H.~Wang, J.~Mao, W.~X. Zhao, C.~Wang, P.~Jiang, and J.-R. Wen,
  ``Feature-aware diversified re-ranking with disentangled representations for
  relevant recommendation,'' in \emph{Proceedings of the 28th ACM SIGKDD
  Conference on Knowledge Discovery and Data Mining}, 2022, pp. 3327--3335.

\bibitem{liu2022rating}
H.~Liu, D.~Tang, J.~Yang, X.~Zhao, H.~Liu, J.~Tang, and Y.~Cheng, ``Rating
  distribution calibration for selection bias mitigation in recommendations,''
  in \emph{Proceedings of the ACM Web Conference 2022}, 2022, pp. 2048--2057.

\bibitem{lu2018like}
Y.~Lu, R.~Dong, and B.~Smyth, ``Why i like it: multi-task learning for
  recommendation and explanation,'' in \emph{Proceedings of the 12th ACM
  Conference on Recommender Systems}, 2018, pp. 4--12.

\bibitem{ma2018modeling}
J.~Ma, Z.~Zhao, X.~Yi, J.~Chen, L.~Hong, and E.~H. Chi, ``Modeling task
  relationships in multi-task learning with multi-gate mixture-of-experts,'' in
  \emph{Proceedings of the 24th ACM SIGKDD international conference on
  knowledge discovery \& data mining}, 2018, pp. 1930--1939.

\bibitem{ma2018entire}
X.~Ma, L.~Zhao, G.~Huang, Z.~Wang, Z.~Hu, X.~Zhu, and K.~Gai, ``Entire space
  multi-task model: An effective approach for estimating post-click conversion
  rate,'' in \emph{The 41st International ACM SIGIR Conference on Research \&
  Development in Information Retrieval}, 2018, pp. 1137--1140.

\bibitem{peska2011upcomp}
L.~Peska, A.~Eckhardt, and P.~Vojtas, ``Upcomp-a php component for
  recommendation based on user behaviour,'' in \emph{2011 IEEE/WIC/ACM
  International Conferences on Web Intelligence and Intelligent Agent
  Technology}, vol.~3.\hskip 1em plus 0.5em minus 0.4em\relax IEEE, 2011, pp.
  306--309.

\bibitem{pradhan2012wish}
S.~Pradhan, P.~R. Krishna, S.~S. Rout, and K.~Jonna, ``Wish-list based shopping
  path discovery and profitable path recommendations,'' in \emph{2012 Third
  International Conference on Services in Emerging Markets}.\hskip 1em plus
  0.5em minus 0.4em\relax IEEE, 2012, pp. 101--106.

\bibitem{richardson2007predicting}
M.~Richardson, E.~Dominowska, and R.~Ragno, ``Predicting clicks: estimating the
  click-through rate for new ads,'' in \emph{Proceedings of the 16th
  international conference on World Wide Web}, 2007, pp. 521--530.

\bibitem{takada2021pop}
R.~Takada, K.~Hoshimure, T.~Iwamoto, and J.~Baba, ``Pop cart: Product
  recommendation system by an agent on a shopping cart,'' in \emph{2021 30th
  IEEE International Conference on Robot \& Human Interactive Communication
  (RO-MAN)}.\hskip 1em plus 0.5em minus 0.4em\relax IEEE, 2021, pp. 59--66.

\bibitem{tang2020progressive}
H.~Tang, J.~Liu, M.~Zhao, and X.~Gong, ``Progressive layered extraction (ple):
  A novel multi-task learning (mtl) model for personalized recommendations,''
  in \emph{Proceedings of the 14th ACM Conference on Recommender Systems},
  2020, pp. 269--278.

\bibitem{vaswani2017attention}
A.~Vaswani, N.~Shazeer, N.~Parmar, J.~Uszkoreit, L.~Jones, A.~N. Gomez,
  {\L}.~Kaiser, and I.~Polosukhin, ``Attention is all you need,''
  \emph{Advances in neural information processing systems}, vol.~30, 2017.

\bibitem{wang2023sequential}
C.~Wang, W.~Ma, C.~Chen, M.~Zhang, Y.~Liu, and S.~Ma, ``Sequential
  recommendation with multiple contrast signals,'' \emph{ACM Transactions on
  Information Systems}, vol.~41, no.~1, pp. 1--27, 2023.

\bibitem{wang2022escm2}
H.~Wang, T.-W. Chang, T.~Liu, J.~Huang, Z.~Chen, C.~Yu, R.~Li, and W.~Chu,
  ``Escm2: Entire space counterfactual multi-task model for post-click
  conversion rate estimation,'' in \emph{Proceedings of the 45th International
  ACM SIGIR Conference on Research and Development in Information Retrieval},
  2022, pp. 363--372.

\bibitem{wang2020next}
Q.~Wang, H.~Yin, T.~Chen, Z.~Huang, H.~Wang, Y.~Zhao, and N.~Q. Viet~Hung,
  ``Next point-of-interest recommendation on resource-constrained mobile
  devices,'' in \emph{Proceedings of the Web conference 2020}, 2020, pp.
  906--916.

\bibitem{wang2021survey}
S.~Wang, L.~Cao, Y.~Wang, Q.~Z. Sheng, M.~A. Orgun, and D.~Lian, ``A survey on
  session-based recommender systems,'' \emph{ACM Computing Surveys (CSUR)},
  vol.~54, no.~7, pp. 1--38, 2021.

\bibitem{wei2021autoheri}
P.~Wei, W.~Zhang, Z.~Xu, S.~Liu, K.-c. Lee, and B.~Zheng, ``Autoheri: Automated
  hierarchical representation integration for post-click conversion rate
  estimation,'' in \emph{Proceedings of the 30th ACM International Conference
  on Information \& Knowledge Management}, 2021, pp. 3528--3532.

\bibitem{wen2021hierarchically}
H.~Wen, J.~Zhang, F.~Lv, W.~Bao, T.~Wang, and Z.~Chen, ``Hierarchically
  modeling micro and macro behaviors via multi-task learning for conversion
  rate prediction,'' in \emph{Proceedings of the 44th International ACM SIGIR
  Conference on Research and Development in Information Retrieval}, 2021, pp.
  2187--2191.

\bibitem{wen2020entire}
H.~Wen, J.~Zhang, Y.~Wang, F.~Lv, W.~Bao, Q.~Lin, and K.~Yang, ``Entire space
  multi-task modeling via post-click behavior decomposition for conversion rate
  prediction,'' in \emph{Proceedings of the 43rd International ACM SIGIR
  Conference on Research and Development in Information Retrieval}, ser. SIGIR
  '20.\hskip 1em plus 0.5em minus 0.4em\relax New York, NY, USA: Association
  for Computing Machinery, 2020, p. 2377–2386.

\bibitem{xu2022amcad}
Z.~Xu, S.~Wen, J.~Wang, G.~Liu, L.~Wang, Z.~Yang, L.~Ding, Y.~Zhang, D.~Zhang,
  J.~Xu, and B.~Zheng, ``Amcad: Adaptive mixed-curvature representation based
  advertisement retrieval system,'' in \emph{2022 IEEE 38th International
  Conference on Data Engineering (ICDE)}, 2022, pp. 3439--3452.

\bibitem{zhang2020large}
W.~Zhang, W.~Bao, X.-Y. Liu, K.~Yang, Q.~Lin, H.~Wen, and R.~Ramezani,
  ``Large-scale causal approaches to debiasing post-click conversion rate
  estimation with multi-task learning,'' in \emph{Proceedings of The Web
  Conference 2020}, 2020, pp. 2775--2781.

\bibitem{zhang2022price}
X.~Zhang, B.~Xu, L.~Yang, C.~Li, F.~Ma, H.~Liu, and H.~Lin, ``Price does
  matter! modeling price and interest preferences in session-based
  recommendation,'' in \emph{Proceedings of the 45th International ACM SIGIR
  Conference on Research and Development in Information Retrieval}, 2022, pp.
  1684--1693.

\bibitem{zhang2022picasso}
Y.~Zhang, L.~Chen, S.~Yang, M.~Yuan, H.~Yi, J.~Zhang, J.~Wang, J.~Dong, Y.~Xu,
  Y.~Song, Y.~Li, D.~Zhang, W.~Lin, L.~Qu, and B.~Zheng, ``Picasso: Unleashing
  the potential of gpu-centric training for wide-and-deep recommender
  systems,'' in \emph{2022 IEEE 38th International Conference on Data
  Engineering (ICDE)}, 2022, pp. 3453--3466.

\bibitem{zhang2022towards}
Z.-Y. Zhang, X.-R. Sheng, Y.~Zhang, B.~Jiang, S.~Han, H.~Deng, and B.~Zheng,
  ``Towards understanding the overfitting phenomenon of deep click-through rate
  models,'' in \emph{Proceedings of the 31st ACM International Conference on
  Information \& Knowledge Management}, 2022, pp. 2671--2680.

\bibitem{zhou2019deep}
G.~Zhou, N.~Mou, Y.~Fan, Q.~Pi, W.~Bian, C.~Zhou, X.~Zhu, and K.~Gai, ``Deep
  interest evolution network for click-through rate prediction,'' in
  \emph{Proceedings of the AAAI conference on artificial intelligence},
  vol.~33, no.~01, 2019, pp. 5941--5948.

\bibitem{zhou2018deep}
G.~Zhou, X.~Zhu, C.~Song, Y.~Fan, H.~Zhu, X.~Ma, Y.~Yan, J.~Jin, H.~Li, and
  K.~Gai, ``Deep interest network for click-through rate prediction,'' in
  \emph{Proceedings of the 24th ACM SIGKDD international conference on
  knowledge discovery \& data mining}, 2018, pp. 1059--1068.

\end{thebibliography}

\end{document}